\documentclass{article}
\usepackage{microtype}
\usepackage{graphicx}
\usepackage{subfigure}
\usepackage{booktabs}
\usepackage{hyperref}

\usepackage[accepted]{icml2025}
\usepackage{amsmath}
\usepackage{amssymb}
\usepackage{mathtools}
\usepackage{amsthm}
\usepackage[capitalize,noabbrev]{cleveref}
\theoremstyle{plain}

\theoremstyle{definition}

\theoremstyle{remark}

\usepackage[textsize=tiny]{todonotes}
\icmltitlerunning{Mechanistic Personality Analysis of LLMs}
\begin{document}
\twocolumn[
\icmltitle{Mechanistic Personality Analysis of LLMs \\[0.2em]
       Steering Personality via Latent Feature Interventions}
\begin{icmlauthorlist}
\icmlauthor{David Courtis}{queen}
\icmlauthor{Ting Hu}{queen}
\end{icmlauthorlist}
\icmlaffiliation{queen}{Department of Computing, Queen's University Kingston, Kingston, ON, Canada}
\icmlcorrespondingauthor{David Courtis}{david.courtis1@gmail.com}
\vskip 0.3in
]
\printAffiliationsAndNotice{}
\begin{abstract}
Large Language Models (LLMs) have demonstrated the ability to simulate human-like OCEAN personality traits in generated text. Previous efforts have focused on prompt engineering or fine-tuning to shape LLM personality. In this work, we propose a mechanistic interpretability approach that directly intervenes on the model's latent features. Our method identifies latent directions in the residual stream corresponding to a target OCEAN trait using sparse autoencoders (SAEs) and contrastive activation analysis. We formalize an \emph{additive steering vector} in activation space and demonstrate how applying a small additive shift to the hidden states enhances the target trait while preserving overall language modeling performance. To determine the optimal combination of feature shifts, we explore a linear weighting heuristic with grid search optimization that balances personality expression with task performance. Our approach shows promise in controllably steering personality traits at the mechanistic level while maintaining high performance on standard benchmarks.
\end{abstract}
\section{Introduction}
Modern large language models (LLMs) are capable of not only generating coherent text but also simulating complex psychological attributes such as personality \citep{serapiogarcia2025personality,sorokovikova2024bigpersonality}. These synthetic personalities emerge as properties of the training process, which relies on vast amounts of human-generated data containing characteristic patterns of thought, feeling, and behavior \citep{wei2022emergentabilitieslargelanguage}. Current work has shown that by designing structured prompts, one can induce distinct personality profiles---for example, extraversion or introversion---that align with psychometric frameworks like the Myers-Briggs Type Indicator (MBTI), IPIP-NEO, or Big Five (OCEAN) \citep{serapiogarcia2025personality} (see Section~\ref{sec:personality} for OCEAN traits breakdown).

While personality shaping has shown to be possible \citep{molchanova2025exploringpotentiallargelanguage, widiger2019}, current methods (prompt engineering or fine-tuning) for personality shaping suffer from inconsistency, resource intensity, and potential side effects such as degraded performance on core language tasks \citep{reynolds2021promptprogramminglargelanguage, fatemi2024comparativeanalysisinstructionfinetuning}. The challenge of concurrently shaping multiple traits is particularly difficult due to their intercorrelated nature, often requiring significant model capacity that smaller or less optimized models may lack \citep{hagendorff2024machinepsychology, lee2025llmsdistinctconsistentpersonality}.

This work introduces a novel framework that directly intervenes on the LLM's internal representations to modulate personality. By leveraging mechanistic interpretability, we decompose the residual activations of the model into interpretable features using sparse autoencoders (SAEs). This approach allows us to identify monosemantic latent features that correlate with high-level personality traits. We then compute an \emph{additive steering vector} that, when applied to the model's hidden states, reliably shifts the output towards the desired personality. The implications of this work extend beyond technical innovation to responsible AI development, as LLM agents have been observed to inadvertently manifest undesirable personality characteristics that can negatively impact safety, fairness, effectiveness in communication, and applications in computational social science and psychology research \citep{wang2025exploringimpactpersonalitytraits, hagendorff2024machinepsychology}.

Our contributions consist of:
\begin{enumerate}
\item We propose a methodology for extracting personality-relevant features via contrastive analysis between positive and negative prompt conditions.
\item We establish a multi-faceted evaluation framework that measures personality expression using embedding-based similarity, LLM-based classification, and human evaluation.
\item We develop refined strategies for computing optimal additive shifts in the latent space: a bidirectional linear weighting approach with grid search optimization.
\end{enumerate}

This approach provides an interpretable and fine-grained method for personality steering between high-level psychometric observations and low-level neural mechanisms.
\section{Literature Review}
\subsection{LLMs and Simulated Personality}
Recent works have demonstrated that LLMs inherently simulate personality traits, which can be extracted via structured prompts and validated against psychological inventories \citep{serapiogarcia2025personality,sorokovikova2024bigpersonality}. In particular, studies have shown that models like GPT-4 and Llama-2 exhibit stable personality profiles when probed with well-designed linguistic cues \citep{PetersMatz2024}. These findings highlight that personality is not simply an emergent artifact but rather a reproducible phenomenon rooted in the training data and model architecture. Research indicates that larger models that have undergone instruction fine-tuning tend to exhibit more pronounced and consistent personality traits, suggesting that model sophistication correlates with the ability to simulate human-like personalities. For instance, Flan-PaLM 540B showed stronger evidence of reliable and valid personality measurements compared to smaller models, which exhibited less control and consistency \citep{serapiogarcia2025personality}.

The analysis of personality through language has a long history in computational linguistics and psychology. Early approaches rely on linguistic cues and psycholinguistic features (e.g., word usage frequencies, sentiment, or syntax preferences) to predict an author's personality traits. With the advent of deep learning, transformer-based models have achieved state-of-the-art results in personality detection from text. For example, \citet{serapiogarcia2025personality} combined a pre-trained BERT model with rich psycholinguistic feature descriptors to improve trait classification accuracy on benchmark datasets (essays and social media posts). Such models implicitly learn that certain patterns of word choice and tone are indicative of personality traits (e.g., frequent social and positive emotion words correlating with high Extraversion) \citet{kerz-etal-2022-pushing}.

\subsection{Personality Shaping Techniques}
Conventional methods for personality shaping include prompt-based interventions and fine-tuning. Prompt-based methods inject personality descriptors directly into the input, but their effectiveness can vary significantly with model size and configuration \citep{serapiogarcia2025personality}. Fine-tuning, while potentially more robust, requires significant computational resources and risks altering other aspects of the model's behavior \citep{AlmuqhimSaeed2021}. Recently, activation-level interventions such as Contrastive Activation Addition (CAA) have shown that directly manipulating internal activations can achieve desired behavioral changes with minimal external modifications \citep{arditi2024refusal}.

In the context of LLMs, recent research has examined whether these models exhibit consistent personality-like behaviors. \citet{serapiogarcia2025personality} conducted extensive evaluations by administering standardized personality questionnaires (based on the Big Five inventory) to a variety of LLMs. Their findings indicate that some models give reliably consistent responses mapping to specific personality profiles, especially when prompted in a way that encourages the model to answer as a human would on a personality test. They also showed that it is possible to direct an LLM to adopt a given personality profile through careful prompting (for instance, asking the model to respond "as if it were an extroverted person"), although the degree of success varies by model size and tuning.

\subsection{Mechanistic Interpretability and Sparse Autoencoders}
Mechanistic interpretability aims to demystify neural network computations by decomposing complex activations into interpretable components. Sparse autoencoders (SAEs) have emerged as a particularly effective tool for this purpose, as they can extract interpretable features from the uninterpretable high-dimensional latent space within a typical LLM layer \citep{cunningham2023interpretable,bricken2023monosemanticity,templeton2024scaling,elhage2022superposition}. These extracted features offer concise, human-understandable representations of semantic concepts. For instance, an SAE may reveal specific directions in the residual stream that correlate with syntactic structures, nouns, or concepts (e.g., apple, sky, red, code bugs, golden gate bridge, sycophantic praise are all potentially individually encoded in this space).

SAEs are specialized neural networks designed to learn efficient data representations by enforcing sparsity on hidden layer activations. This sparsity constraint ensures that most hidden units remain inactive, resulting in compact and meaningful feature representations. They are particularly valuable for addressing the challenge of \emph{superposition} in large models, where the number of latent concepts a model encodes can far exceed the number of neurons, forcing neurons to multiplex multiple concepts, a concept named \emph{polysemanticity}. To cut through this complexity, researchers have employed \emph{sparse dictionary learning} on model activations, seeking an overcomplete basis in which each basis element (feature) corresponds to a more \emph{monosemantic concept} \citep{bricken2023monosemanticity}.

In \emph{Towards Monosemanticity}, Bricken et al. applied an SAE to the activations of a transformer, discovering interpretable features that captured distinct aspects of the input or context (e.g., a feature that activates on legal or formal language, another on lines of computer code). Building on this, \citet{templeton2024scaling} scaled the approach to a larger model and demonstrated that many of these features carry semantic meaning and can be manipulated individually. In their case studies, toggling a single feature's activation (by inserting a targeted activation vector into the model's hidden state) led to qualitatively noticeable changes in the model's output focused on that feature's concept, all while other aspects of the output remained coherent.

Together, these advances provide the foundation for our approach: by extracting and controlling interpretable latent features, we introduce a novel and reliable personality steering method aimed at addressing the drawbacks of conventional approaches.
\section{Methodology}
\label{sec:methodology}
\begin{figure}[h]
\centering
\includegraphics[width=0.9\linewidth]{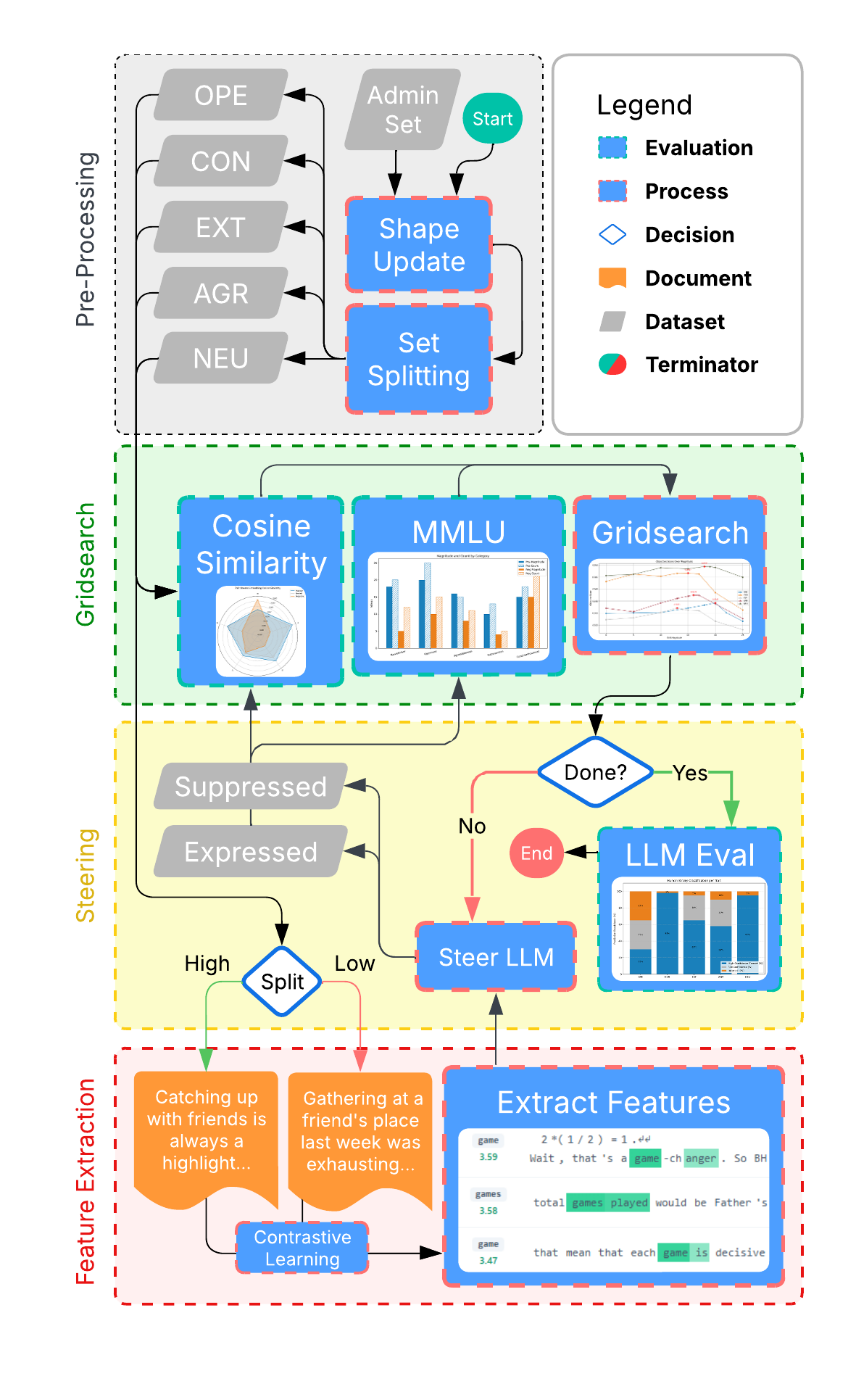}
\caption{Overview of our personality steering pipeline, showing the flow from data generation through feature extraction, optimization, and evaluation. Processes transform the data into results, while evaluations transform results into data for the algorithm and results analysis. The gridsearch algorithm involves two benchmarks with the third being performed post-gridsearch. Each iteration of the gridsearch algorithm involves a modification to the LLM steering parameters of the operation.}
\label{fig:pipeline}
\end{figure}

Our methodology combines contrastive feature analysis with activation steering to induce a target personality trait in an LLM. The approach is divided into four main components: Data generation and dataset creation (pre-processing), (2) Extracting personality-relevant features (feature extraction), (3) Behavior benchmarking gridsearch (gridsearch), and (4) Additive feature shift and steering (steering). Figure \ref{fig:pipeline} illustrates our complete pipeline.
\subsection{Data Generation and Dataset Creation}
Our pipeline begins with generating personality-shaped text samples that will serve as the foundation for identifying relevant latent features.

\subsubsection{Status Update Generator with Shaping}
We implement a personality-shaping methodology based on \citet{serapiogarcia2025personality} using the PsyBORGS open source framework to generate Facebook-style status updates that express varying levels of the Big Five personality traits. This approach uses structured prompting to control the intensity of trait expression in model outputs. We apply this to the DeepSeek-R1-Distill-Llama-8B model, producing a set of 12k updates with varying levels of OCEAN trait expression. See Table \ref{tab:shaped_examples} for examples.

\begin{table*}[h]
\caption{Examples of shaped status updates for each trait (high vs. low expression)}
\label{tab:shaped_examples}
\centering
\begin{tabular}{p{1cm}|p{6.5cm}|p{6.5cm}}
\toprule
\textbf{Trait} & \textbf{High Expression} & \textbf{Low Expression} \\
\midrule
CON & I just finished a book on time management. It’s been a game changer for my productivity. & I broke my diet again. Maybe I should start fresh tomorrow. \\
NEU & I just want to scream. This week is a nightmare. & rushing on this playlist someone shared. So many vibez! \\
\bottomrule
\end{tabular}
\end{table*}

\subsubsection{Dataset Creation for Mechanistic Interpretability}
Rather than using a positive and neutral set as in some prior work, we create distinct positive and negative sets for each trait. This approach allows us to:
\begin{itemize}
    \item Identify both trait-enhancing and trait-suppressing features
    \item Filter out common features that appear in both sets
    \item Create a more robust contrast for feature extraction
\end{itemize}

For each trait, we categorize status updates based on their trait expression scores:
\begin{itemize}
    \item \textbf{Negative Set}: Updates with scores 1-3 on a 9-point scale
    \item \textbf{Positive Set}: Updates with scores 7-9 on a 9-point scale
\end{itemize}

For neuroticism, the scoring is reversed, as higher neuroticism corresponds to more negative emotional expression.

\subsection{Identifying Personality-Relevant Features}
\subsubsection{Positive/Negative Prompt Construction}
For a target trait (e.g., extraversion), we construct two prompt sets:
\begin{itemize}
    \item \textbf{Positive Set} \(P\): Prompts designed to elicit high levels of the trait. For instance: \emph{You are an outgoing, energetic individual. Answer in a lively and sociable tone.''}
    \item \textbf{Negative Set} \(N\): Prompts that encourage low trait expression, e.g., \emph{You are a reserved and introspective individual. Respond in a calm and concise manner.''}
\end{itemize}
These prompts are derived from psycholinguistic markers discussed in \citep{serapiogarcia2025personality} and refined using structured prompting techniques.

\subsubsection{Activation Extraction and Contrastive Analysis}
For each prompt \(x \in P \cup N\), we perform a forward pass through the LLM and extract the activation \(h(x) \in \mathbb{R}^d\) from a designated layer. For multi-token inputs, the activation of the final token is used. The mean activations for each set are computed as:
\begin{equation}
\mu_P = \frac{1}{|P|}\sum_{x\in P} h(x), \quad \mu_N = \frac{1}{|N|}\sum_{x\in N} h(x).
\label{eq:mean_activations}
\end{equation}
The raw contrastive activation vector is then defined as:
\begin{equation}
\Delta h = \mu_P - \mu_N.
\label{eq:contrast_vector}
\end{equation}
This vector captures the direction in activation space that corresponds to increased trait expression.

\subsubsection{Sparse Autoencoder (SAE) Decomposition and Feature Ranking}
In our work, we utilize a pretrained sparse autoencoder from qresearch/DeepSeek-R1-Distill-Llama-8B-SAE-l19, which was trained on the LMSYS-Chat-1M dataset specifically for layer 19 of DeepSeek-R1-Distill-Llama-8B. We attach our SAE to layer 19 of 30 (63 \% depth) because mid–late residual-stream blocks have repeatedly been shown to concentrate the most interpretable, semantically rich features while remaining sufficiently upstream of the final logits to avoid heavy output–entanglement. Empirical sweeps on GPT-2-small \citep{gao2024scalingevaluatingsparseautoencoders} and Claude-3 Sonnet \citep{templeton2024scaling} both locate their “sweet-spot’’ in the 60–70 \% depth range; earlier layers mainly encode surface form, whereas the final blocks over-specialise for next-token prediction. The publicly released checkpoint from qresearch therefore offers a near-optimal trade-off—achieving 93 \% \(L_0\) sparsity and low reconstruction error—without the added compute of training a new SAE at neighbouring layers.

As part of our pipeline, we implement a Feature Decomposition component that ranks both positive and negative features by their relevance to each personality trait. This system:
\begin{itemize}
    \item Processes the activations from our positive and negative datasets
    \item Computes mean activation differences between the sets
    \item Decomposes these differences using the pretrained SAE
    \item Ranks features by their significance to trait expression
    \item Saves the ranked features as JSON files for subsequent intervention steps
\end{itemize}

Let \(E: \mathbb{R}^d \rightarrow \mathbb{R}^m\) and \(D: \mathbb{R}^m \rightarrow \mathbb{R}^d\) denote the encoder and decoder of our pretrained SAE, respectively. The SAE enforces sparsity in the latent representation \(z = E(h)\), where \(m \gg d\), such that most components of \(z\) are near zero. Mapping the mean activations yields:
\begin{equation}
\bar{z}_P = E(\mu_P), \quad \bar{z}_N = E(\mu_N),
\label{eq:latent_mappings}
\end{equation}
and the latent difference is:
\begin{equation}
\Delta z = \bar{z}_P - \bar{z}_N.
\label{eq:latent_diff}
\end{equation}
We rank each latent component by the absolute difference \(|\Delta z_i|\) and select the top-\(k_{\text{pos}}\) feature indices \(i_1, i_2, \ldots, i_{k_{\text{pos}}}\) corresponding to the highest differences in the positive direction. Additionally, we invert the positive and negative sets to identify top-\(k_{\text{neg}}\) feature indices \(j_1, j_2, \ldots, j_{k_{\text{neg}}}\) corresponding to features that suppress the trait. This bidirectional approach allows for more nuanced control of personality traits.

For each selected index, we define feature vectors \(w_{i_j}\) and \(w_{j_l}\) as:
\begin{equation}
w_{i_j} = D(e_{i_j}), \quad w_{j_l} = D(e_{j_l}),
\label{eq:feature_vectors}
\end{equation}
where \(e_{i_j}\) and \(e_{j_l}\) are one-hot vectors in \(\mathbb{R}^m\) with a 1 in position \(i_j\) and \(j_l\), respectively. The sets of feature vectors relevant to a personality trait are then:
\begin{equation}
F_{\text{trait}}^{\text{pos}} = \{ w_{i_1}, w_{i_2}, \dots, w_{i_{k_{\text{pos}}}} \}
\label{eq:pos_features}
\end{equation}
\begin{equation}
F_{\text{trait}}^{\text{neg}} = \{ w_{j_1}, w_{j_2}, \dots, w_{j_{k_{\text{neg}}}} \}
\label{eq:neg_features}
\end{equation}

For computational convenience, we normalize each feature vector to create unit vectors:
\begin{equation}
u_{i_j} = \frac{w_{i_j}}{\|w_{i_j}\|}, \quad u_{j_l} = \frac{w_{j_l}}{\|w_{j_l}\|}
\label{eq:unit_vectors}
\end{equation}

\subsection{Behavior Benchmarking and Evaluation Metrics}
\label{sec:personality}
\subsubsection{Personality Expression Measurement}
Our approach to measuring personality expression employs a multi-faceted evaluation framework consisting of three complementary methods:

\paragraph{1. Embedded Cosine Similarity}
We evaluate the similarity between generated text and reference text exemplifying each personality trait using embedding-based comparison. This component of our pipeline integrates with our Status Update Generator with Feature Intervention and Grid Search system, which:
\begin{itemize}
    \item Takes feature JSON files from the Feature Decomposition step
    \item Generates status updates using a set of 8 parameters to steer the model
    \item Evaluates generated outputs to find optimal parameter values
\end{itemize}

The embedding-based evaluation procedure is as follows:

\begin{itemize}
    \item For each trait (Openness, Conscientiousness, Extraversion, Agreeableness, Neuroticism), we maintain reference texts in files (s1.txt, s2.txt) that exemplify high expression of that trait.
    \item We generate output texts under different conditions: trait-enhanced (a.txt), neutral (b.txt), trait-suppressed (c.txt), and baseline (baseline.txt).
    \item We also evaluate random word passages (with practically zero correlation to any trait) to establish a lower bound for our similarity metrics.
    \item Text embeddings are obtained using the OpenAI Embeddings API (text-embedding-ada-002 model).
    \item For each trait, we compute the average embedding of the reference texts and measure cosine similarity between this reference embedding and the embeddings of our generated outputs:
    
    \begin{equation}
    \text{score}_{\text{trait}}(text) = \cos(\text{emb}_{\text{trait}}^{\text{ref}}, \text{emb}_{\text{text}})
    \label{eq:cosine_score}
    \end{equation}
    
    where $\text{emb}_{\text{trait}}^{\text{ref}}$ is the average embedding of reference texts for the trait, and $\text{emb}_{\text{text}}$ is the embedding of the generated text.
\end{itemize}

This method provides a continuous measure of similarity to trait-specific references, allowing for fine-grained comparison of different steering approaches. The evaluation of baseline passages establishes the correlation of basic coherent text with each trait, while the random word passages serve as a control group.

\paragraph{2. LLM-Based Evaluation}
We supplement the embedding-based approach with evaluations from state-of-the-art LLMs (Claude 3.7 and GPT-4o):

\begin{itemize}
    \item We generate 20 paired samples, each consisting of a trait-tuned response and an untuned response.
    \item The samples are presented to the LLMs with instructions to identify which text exhibits higher levels of the target trait.
    \item The prompt follows this format: "You are an objective judge of [Trait Name] in the following Journal post. [Trait Name] relates to [Trait Description]. Out of the provided posts, which post, A or B demonstrates more [Trait Name]? Provide your final Answer in the format Answer: A/B"
    \item For each trait, we use standardized descriptions:
    \begin{itemize}
        \item Openness: cognitive flexibility, curiosity about ideas, reflective thought, engagement with abstract concepts, preference for varied experiences
        \item Conscientiousness: orderliness, systematic planning, self-regulation, dependability, attention to detail, work-related experiences
        \item Extraversion: social engagement, assertiveness, level of interpersonal activity, energetic expression, preference for external stimulation
        \item Agreeableness: cooperativeness, approach to conflict resolution, interpersonal sensitivity, trust propensity, inclination toward social harmony
        \item Neuroticism: emotional reactivity, mood variability, sensitivity to stress, affective instability, responsiveness to negative stimuli
    \end{itemize}
    \item We conduct 200 trials of 20 evaluations each per trait.
    \item A trait modification is considered correct if LLMs correctly identify the trait-tuned response \(>\) 80\% of the time, incorrect if they identify it correctly \(<\) 20\% of the time, and "mid confidence" if the identification rate falls between these thresholds.
\end{itemize}

\paragraph{3. Human Evaluation}
To validate the LLM evaluations, we conduct parallel human evaluations with the same process as LLM evaluation.

Combining quantitative similarity metrics with LLM judgment and human perception provides a more robust assessment of our personality steering technique.

\subsubsection{General Performance Metrics}
To ensure that personality steering does not degrade overall language modeling capabilities, we evaluate:
\begin{itemize}
    \item \textbf{Cosine similarity} of generated text compared to an incoherent word file and a baseline generation (discussed later).
    \item \textbf{MMLU Accuracy} on the Massive Multitask Language Understanding benchmark \citep{jiang2024persona}.
\end{itemize}
These metrics provide a quantitative measure of how the intervention affects both stylistic and functional aspects of the model.

\subsection{Optimal Additive Feature Shifts}
Given the baseline activation \(h\), we aim to modify it to a personality-shifted activation \(h'\) by adding a weighted combination of both positive and negative trait-relevant feature vectors as follows:
\begin{equation}
h' = h + \sum_{j=1}^{k_{\text{pos}}} \delta_{i_j}^{\text{pos}} \, u_{i_j} + \sum_{l=1}^{k_{\text{neg}}} \delta_{j_l}^{\text{neg}} \, u_{j_l},
\label{eq:activation_shift}
\end{equation}
where \(\delta_{i_j}^{\text{pos}}\) and \(\delta_{j_l}^{\text{neg}}\) are scalar coefficients that control the magnitude of the shift along each normalized feature vector.

\subsubsection{Bidirectional Linear Weighting}
Our linear weighting approach now incorporates both positive and negative features:
\begin{equation}
\delta_{i_j}^{\text{pos}} = \alpha_{\text{pos}} \cdot |\Delta z_{i_j}|
\label{eq:pos_weight}
\end{equation}
\begin{equation}
\delta_{j_l}^{\text{neg}} = \alpha_{\text{neg}} \cdot |\Delta z_{j_l}|
\label{eq:neg_weight}
\end{equation}
with global scaling factors \(\alpha_{\text{pos}} > 0\) and \(\alpha_{\text{neg}} < 0\) tuned empirically. This approach allows for both enhancing trait-positive features and suppressing trait-negative features, creating a more balanced and effective personality shift.

\subsubsection{Grid Search Optimization}
We implement a grid search optimization as part of our Status Update Generator with Feature Intervention system. This component takes the ranked features from our Feature Decomposition step and systematically explores the parameter space to find optimal steering configurations.

To find the best combination of feature shifts for personality steering, we maximize the following \emph{objective function}:
\begin{equation}
\begin{split}
\mathcal{O}\bigl(\alpha_{\text{pos}}, k_{\text{pos}}, \alpha_{\text{neg}}, k_{\text{neg}}) = & S_{\text{trait}}(\alpha_{\text{pos}}, k_{\text{pos}}, \alpha_{\text{neg}}, k_{\text{neg}}) \\
&- \lambda \cdot C(\alpha_{\text{pos}}, k_{\text{pos}}, \alpha_{\text{neg}}, k_{\text{neg}}),
\end{split}
\label{eq:objective_function}
\end{equation}
where:
\begin{itemize}
    \item \(S_{\text{trait}}(\cdot)\) is a measure of how strongly the target personality trait is expressed in the model’s generated text,
    \item \(C(\cdot)\) is a penalty term capturing performance degradation. The penalty term is calculated as the relative difference from the baseline.
    \item \(\lambda\) is a scalar weight balancing trait expression against overall performance. 
\end{itemize}
Throughout our experiments, we fix \(\lambda = 0.5\). In other words, we set equal importance on increasing the personality expression score and on limiting performance degradation. The grid search proceeds by systematically varying 
\(\alpha_{\text{pos}}, k_{\text{pos}}, \alpha_{\text{neg}},\) and 
\(k_{\text{neg}}\), then choosing the parameter set that maximizes 
\(\mathcal{O}(\cdot)\).

The grid search proceeds in multiple phases:
\begin{enumerate}
    \item Initial coarse search: We perform a 5×5×5×5 grid search over ranges of 0-25 for both shift magnitudes (\(\alpha_{\text{pos}}\) and \(\alpha_{\text{neg}}\)) and for both feature counts (\(k_{\text{pos}}\) and \(k_{\text{neg}}\)).
    \item Refinement: Once an optimal point in the 4D grid is identified, we refine each dimension individually while holding the others constant.
    \item Granular search: We first explore neighboring points at ±2 units from the current optimum. If a better value is found, we then search the midpoint between that value and the original optimum, iterating until convergence.
\end{enumerate}

For each configuration in the grid, our system:
\begin{itemize}
    \item Generates status updates with the specified feature interventions
    \item Evaluates personality expression using embedding cosine similarity
    \item Measures model performance on MMLU to ensure cognitive abilities remain intact
    \item Calculates the combined objective function value
\end{itemize}

This progressive refinement approach allows us to efficiently explore the high-dimensional parameter space and identify the most effective personality steering configuration while balancing trait expression and model performance.

\subsubsection{Steering Implementation}
With the optimal parameters determined, we apply the intervention during inference at the target layer:
\begin{equation}
h' = h + \sum_{j=1}^{k_{\text{pos}}^*} \alpha_{\text{pos}}^* \cdot |\Delta z_{i_j}| \, u_{i_j} + \sum_{l=1}^{k_{\text{neg}}^*} \alpha_{\text{neg}}^* \cdot |\Delta z_{j_l}| \, u_{j_l}.
\label{eq:final_intervention}
\end{equation}
This modification is applied without altering the underlying model parameters, allowing for on-demand personality modulation that can be easily enabled or disabled.

\subsection{Latent Feature Intervention Experiment}
To establish causality between identified features and personality traits, we perform controlled intervention experiments. The procedure is as follows: for a given personality trait $T$ (for example, Extraversion, Agreeableness, or another Big Five trait) and a specific feature vector $w_{i_j}$ that we hypothesize to be related to trait $T$, we modify the model's activation on new test inputs to simulate enhancing or suppressing that feature and observe the resulting effect on the expression of trait $T$ in the model's output.

During the model's text generation, at the point where we compute the layer-$\ell$ activation $h = f_\ell(x)$ for the current context, we intervene by altering the activation vector before it continues through the remaining layers. Specifically, we adjust the corresponding feature coefficient in the sparse representation. We obtain the sparse code $a = a(x)$ for the activation $h$ (using the same encoder from the dictionary learning step). Then:

\begin{itemize}
    \item To \textbf{activate} feature vector $w_{i_j}$, we increase the corresponding coefficient $a_{i_j}$ by some amount $\Delta$ (adding a positive offset), enhancing that feature's influence on the model's computation.
    \item To \textbf{suppress} feature vector $w_{i_j}$, we decrease coefficient $a_{i_j}$ or set it to zero (if it was positive) or to a large negative value (if we allow features to take negative activations in the code), reducing or eliminating that feature's influence.
\end{itemize}

We then reconstruct a modified activation $\tilde{h} = W^\top a$ using the altered code. This modified activation $\tilde{h}$ replaces the original activation $h$ for layer $\ell$ in the forward pass of the model. The rest of the model then generates the continuation based on this intervened activation.

We apply this procedure to multiple test prompts, comparing the model's outputs under three experimental conditions: \emph{baseline} (no intervention), \emph{feature-activated}, and \emph{feature-suppressed}. For fairness, all other aspects of generation (sampling strategy, random seed, etc.) are kept the same across conditions. We then measure the personality trait scores of the outputs in each condition using the evaluation methods described in Section 3.2.1.

\subsection{Model Implementation Details}
We conduct our experiments on DeepSeek-R1-Distill-Llama-8B, an 8B-parameter LLM architecture at full precision; however, this experiment may be conducted similarly on other architectures. The layer $\ell$ targeted for feature extraction is the output of the 19th transformer block's feedforward sublayer. This layer has dimensionality $d=4096$ neurons.

We utilize the pretrained sparse autoencoder from qresearch/DeepSeek-R1-Distill-Llama-8B-SAE-l19, which was trained on the LMSYS-Chat-1M dataset. This SAE has a feature dimensionality $m = 8d = 32768$, providing an 8-fold overcomplete representation, consistent with prior work that found features in this range to be effective for capturing monosemantic concepts \citep{bricken2023monosemanticity,templeton2024scaling}.

For our evaluation, we generated 200 new prompts covering a variety of everyday and knowledge topics (without any personality-specific instruction) and collected the baseline outputs from our model. These outputs are typically a few sentences long each. We recorded the activation codes $a(x)$ for each prompt at the beginning of generation and computed personality trait scores for each output using our multi-faceted evaluation approach. 

From this data, for each personality trait $T$ (e.g., Extraversion, Conscientiousness, etc.), we identified both the positive feature vectors $w_{i^*}$ that showed the highest correlation with increased trait expression, and the negative feature vectors $w_{j^*}$ that showed the highest correlation with decreased trait expression. These identified feature vectors serve as our primary candidates for personality-relevant features in both positive and negative directions.

We then performed interventions on these selected feature vectors for each trait as described in Section 3.4. The intervention strength parameters were calibrated through our grid search optimization to find the optimal balance between personality expression and model performance.
\section{Results and Discussion}
\label{sec:results}

\subsection{Steering Gives Expected Results}
\begin{figure}[t]
\centering
\includegraphics[width=0.9\linewidth]{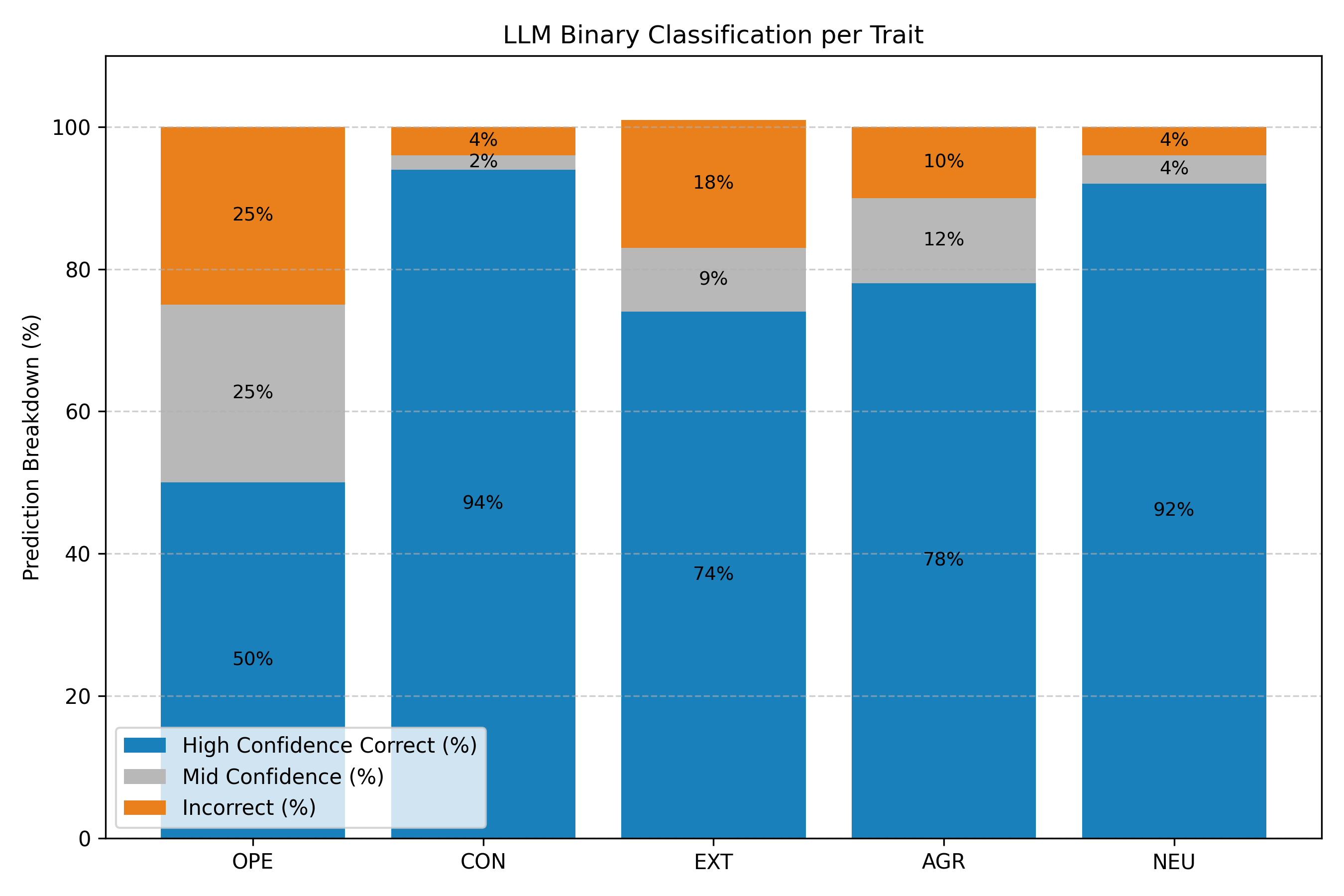}
\caption{LLM Binary Classification per Trait, showing the percentage of correct (high confidence), low confidence, and incorrect classifications for each OCEAN trait. Conscientiousness and Neuroticism show the highest detection rates \(>90\%\), while Openness presents significant challenges.}
\label{fig:llm_classification}
\end{figure}

\begin{figure}[t]
\centering
\includegraphics[width=0.9\linewidth]{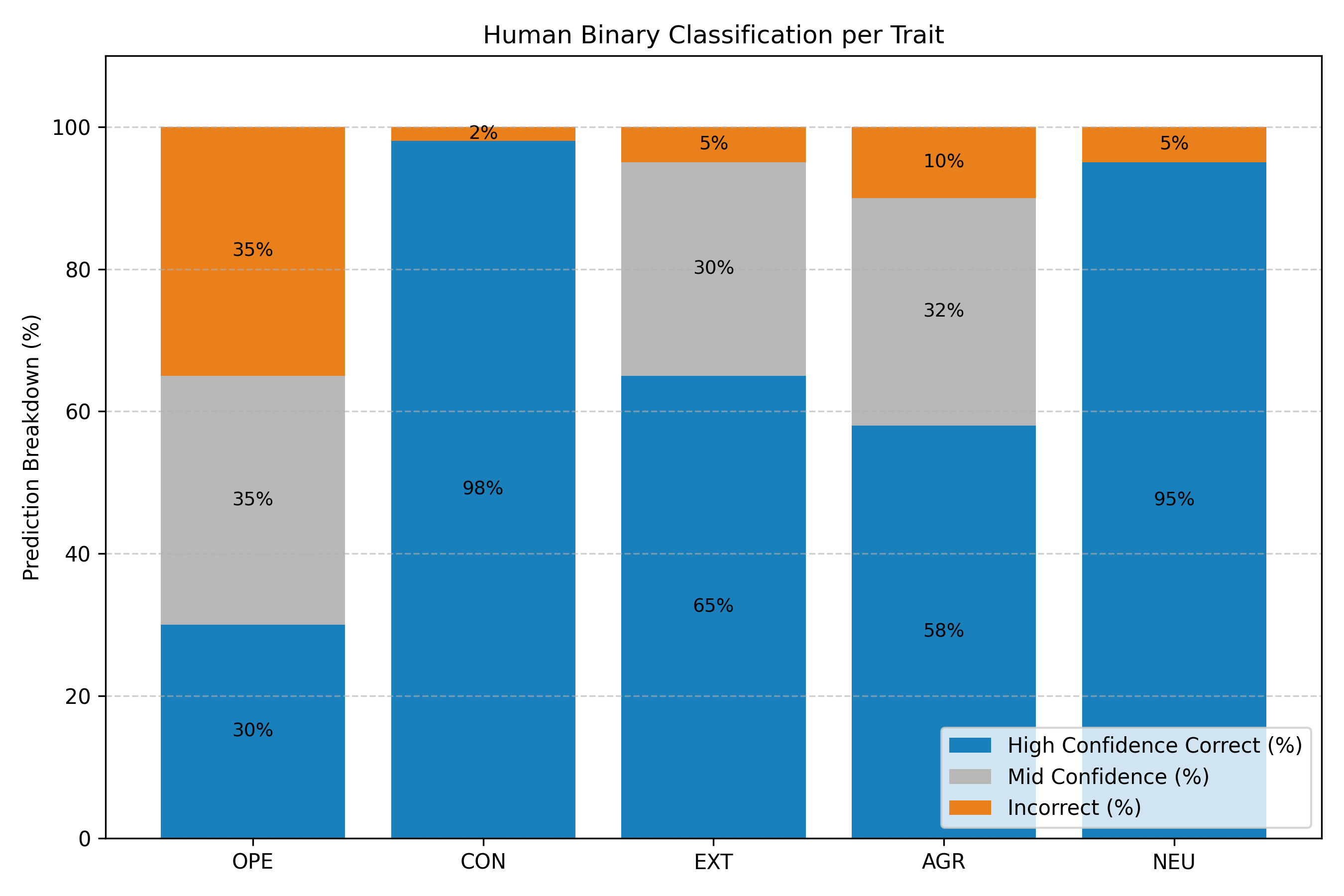}
\caption{Human Binary Classification per Trait, showing the percentage of correct (high confidence), low confidence, and incorrect classifications for each OCEAN trait. Note the similar pattern to LLM classification, with higher uncertainty for Openness and Agreeableness.}
\label{fig:human_classification}
\end{figure}

\begin{figure}[t]
\centering
\includegraphics[width=0.9\linewidth]{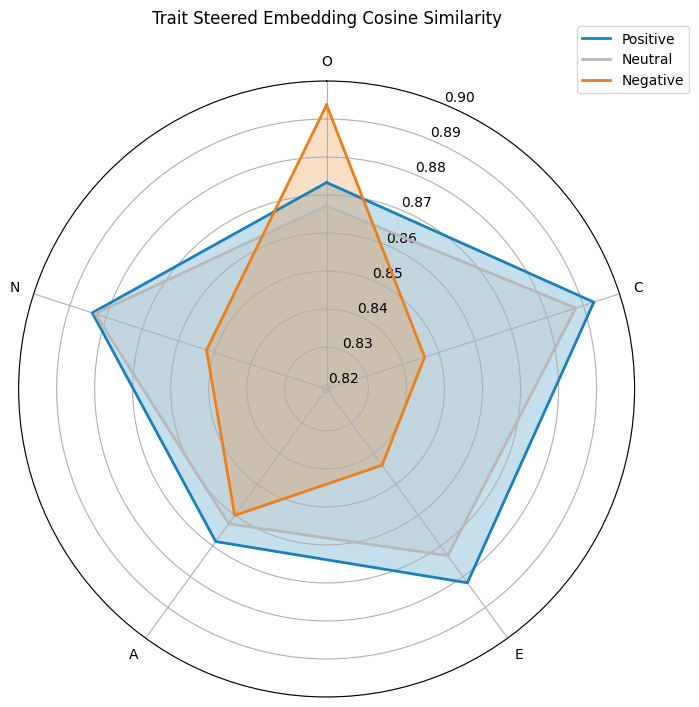}
\caption{Trait Embedding Cosine Similarity between reference trait descriptions and generated text under baseline (yellow), positive shift (orange), and negative shift (pink) conditions. The radar chart illustrates how our interventions consistently amplify positive trait expressions while generally reducing negative trait expressions.}
\label{fig:cosine_similarity}
\end{figure}

Our multi-faceted evaluation approach reveals significant insights into the effectiveness of latent feature interventions for personality shaping. Figures \ref{fig:llm_classification} and \ref{fig:human_classification} present the binary classification results from LLMs and human evaluators, while Figure \ref{fig:cosine_similarity} displays the embedding-based cosine similarity across traits.

The classification results reveal a clear pattern: both LLMs and humans could reliably detect our personality interventions for most traits, but with notable variations. Conscientiousness (94\% LLM, 98\% human correct) and Neuroticism (92\% LLM, 95\% human correct) showed the highest detection rates, suggesting these traits manifest in particularly clear linguistic patterns that are readily identifiable. Extraversion (74\% LLM, 65\% human correct) and Agreeableness (78\% LLM, 58\% human correct) showed moderate detection rates, while Openness proved challenging (50\% LLM, 30\% human correct).

This difficulty in shaping Openness is particularly intriguing. One possible explanation is that large language models are inherently "open" by design—they are trained to generate diverse, creative, and intellectually curious responses \citet{jiang2024persona}. As \citet{hilliard2024elicitingpersonalitytraitslarge} noted, LLMs often score naturally high on Openness measures, making further enhancement less distinctive. Alternatively, the concept of Openness may manifest in more subtle linguistic patterns that are harder to detect through our intervention method.

The cosine similarity measurements provide additional evidence for the effectiveness of our approach. Positive shifts consistently increased similarity to trait reference embeddings across all traits, with the most dramatic increases observed for Openness and Extraversion. Interestingly, our negative shifts produced mixed results: dramatically decreasing similarity for conscientiousness, extraversion, and neuroticism, while having minimal impact on agreeableness.

The unusual pattern observed for Openness, where negative shifts produced higher similarity than positive shifts, suggests a complex relationship between feature activation and trait expression. This counterintuitive finding might indicate that, for openness, the absence of certain concepts—rather than their presence—most strongly signals the trait. As observed in \citet{PetersMatz2024}, personality traits can manifest through both the commission and omission of certain linguistic patterns.

\begin{figure*}[t]
\centering
\includegraphics[width=\textwidth]{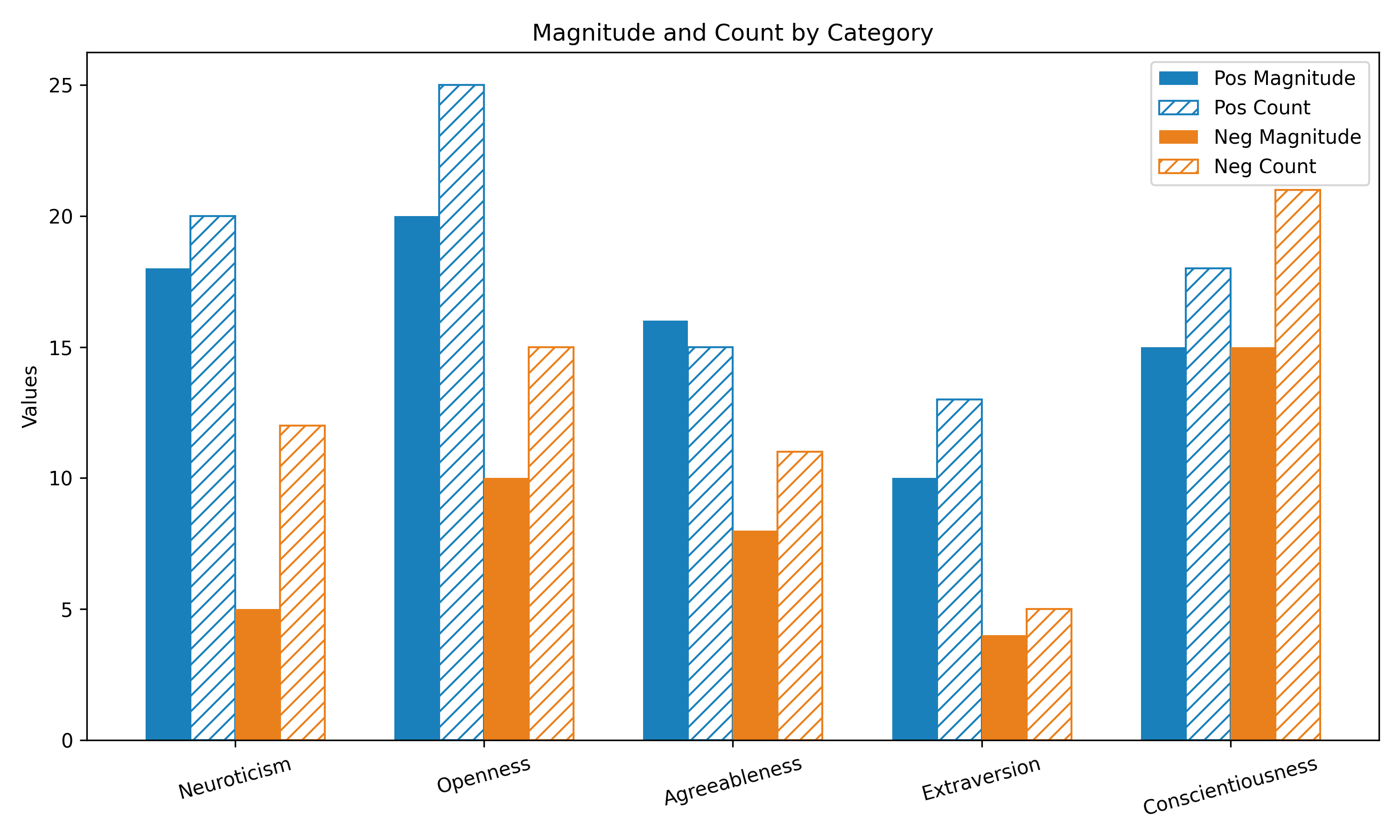}
\caption{Optimal parameters identified through grid search for each OCEAN trait, showing positive and negative feature magnitudes and counts. Note how Openness and Conscientiousness require the most features, while Extraversion requires the fewest.}
\label{fig:gridsearch_optimums}
\end{figure*}

We acknowledge that cosine similarity alone (Equation \ref{eq:cosine_score}) provides an incomplete picture. The embedding space may conflate the conceptual discussion of a trait with the actual expression of that trait: Discussing openness is not the same as demonstrating it. This supports the importance of our multifaceted evaluation approach.

\subsection{Gridsearch Exploration and Optimums}
\begin{figure*}[t]
\centering
\includegraphics[width=\textwidth]{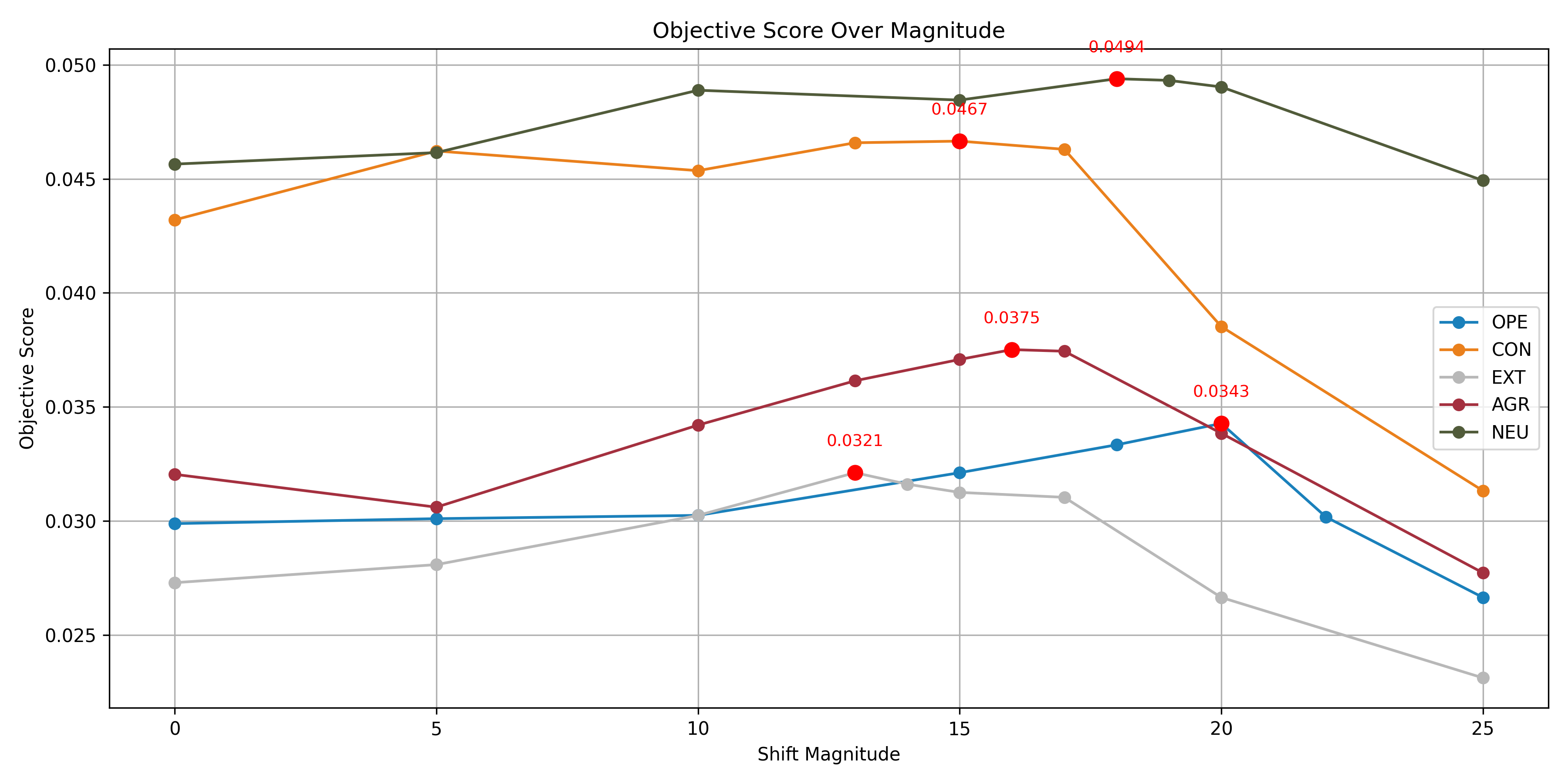}
\caption{Objective score as a function of shift magnitude for each OCEAN trait, highlighting the optimal magnitude (marked with red) before performance degradation. Neuroticism and Openness show the highest peak scores, while all traits exhibit diminishing returns with excessive intervention.}
\label{fig:objective_score}
\end{figure*}

Our intervention optimization revealed the optimal parameters for each personality trait, as shown in Figures~\ref{fig:gridsearch_optimums} and \ref{fig:objective_score}. 
Recall that the objective function we optimize is given by:
\begin{equation}
\mathcal{O}\bigl(\alpha_{\text{pos}}, k_{\text{pos}}, \alpha_{\text{neg}}, k_{\text{neg}} \bigr) 
= S_{\text{trait}}(\cdot) 
+ \lambda \cdot C(\cdot),
\end{equation}
where $S_{\text{trait}}$ measures how strongly the target trait is expressed, and $C(\cdot)$ is a penalty term calculated as the relative difference from the baseline MMLU score.

In Figure~\ref{fig:objective_score}, we plot the combined objective $\mathcal{O}(\cdot)$ as a function of the magnitude of the shift for each trait.

All traits exhibit an initial gain in $\mathcal{O}(\cdot)$ as the magnitude of the shift increases, followed by a clear peak and subsequent degradation. This underscores the trade-off between enhancing personality expression and preserving overall language-model performance. We discovered through direct observation of the generated response that the steep dropoff is usually due to partial or complete loss of coherency, resulting in poor MMLU and embedding scores.

The optimal parameters vary substantially across traits, revealing important differences in how these personality dimensions are encoded in the model. Openness required the highest number of positive features (25) and a relatively high positive magnitude (20), suggesting that this trait is distributed across many latent features requiring significant amplification. Conscientiousness showed a more balanced profile, with moderate counts for both positive (18) and negative (21) features at equal magnitudes (15), indicating that this trait requires both enhancement of certain features and suppression of others for optimal expression.

Perhaps most striking is Extraversion, which required the fewest features overall (9 positive, 4 negative) at moderate magnitudes (13 positive, 5 negative). This parsimony suggests that Extraversion may be encoded in a more concentrated set of latent features, making it more amenable to targeted intervention. Indeed, \citet{AlmuqhimSaeed2021} noted that Extraversion tends to have more consistent and recognizable linguistic markers than other traits, which aligns with our finding of a more compact feature representation.

The objective score trajectories provide further insights into the relationship between feature intervention and trait expression. All traits showed an initial increase in objective score as shift magnitude increased, followed by a performance decline past an optimal point. This pattern of diminishing returns suggests a delicate balance—while some feature amplification enhances trait expression, excessive intervention disrupts the model's underlying language capabilities.

The optimal shift magnitude varied across traits: Neuroticism peaked at approximately 18-20, followed by Openness near 20, while Conscientiousness, Agreeableness, and Extraversion showed earlier peaks around 15. This difference in tolerance for intervention magnitude may reflect varying degrees of feature redundancy or plasticity across trait dimensions. Traits with higher optimal magnitudes may have more distributed representations that can accommodate larger shifts without degrading coherence.

Interestingly, our analysis of the grid search optimums (Figure \ref{fig:gridsearch_optimums}) reveals that traits requiring more features also tended to sustain higher magnitudes before degradation (Figure \ref{fig:objective_score}). This correlation suggests that trait complexity—as measured by the number of relevant features—may predict tolerance for intervention intensity. More complex traits distributed across many features appear more robust to feature manipulation, possibly because the intervention affects a smaller percentage of each feature's overall contribution to the model's language generation.

These findings align with psychological research suggesting that traits like Openness and Neuroticism encompass more varied facets than traits like Extraversion \citep{serapiogarcia2025personality}. Our results provide a novel computational perspective on this facet structure, suggesting that the distributed nature of certain traits in LLMs may mirror their psychological complexity in humans; however, more work on the composite features needs to be done to fully establish that conjecture.

\subsection{Empirical Evaluation of Personality Steering}

Table~\ref{tab:tuned_short} shows a concise excerpt of our personality steering results for three of the Big Five traits (Openness, Extraversion, and Neuroticism). We present only short excerpts here due to space constraints. The full table for all five traits can be found in Appendix~\ref{app:full_table} (Table~\ref{tab:tuned_comparison_full}). More results can be found as well within the results section of our codebase.
\begin{quote}
\small
"Generate a long and detailed journal update as a person describing what they have been up to recently. This should give a broad and varied view of their experience, either positive or negative. Talk about a wide variety of things."
\end{quote}

\begin{table*}[ht]
\centering
\caption{Comparison of Tuned Up vs. Tuned Down Responses for Personality Traits}
\label{tab:tuned_short}
\begin{tabular}{|p{1cm}|p{6.5cm}|p{6.5cm}|}
\hline
\textbf{Trait} & \textbf{Tuned Up Response Excerpt} & \textbf{Tuned Down Response Excerpt} \\
\hline
OPE & One of the most transformative things I’ve done lately is diving into a new hobby ... I’ve pushed myself beyond my comfort zones, learning new techniques, and embracing the chaos of creativity.... & The days have been a bit of a mix lately—some moments of pure joy, some moments of total chaos, and a few that have left me wondering where the time goes.... \\
\hline
CON & ....There’s something about sitting here with a cup of coffee, staring at these spreadsheets and spreadsheets, like it’s some kind of therapy.... & My free time is a joke. I always end up scrolling through social media until 2 a.m....Then I marathon-read for hours.... \\
\hline
EXT & The blog is one of my main hobbies.... It's a creative outlet that I’m really enjoying .... It's like creating my own world in the digital space....  & I've been spending so much time with this friend lately, and I'm just so tired, I can't even think about talking to my family.... \\
\hline
AGR & ....volunteering at a local charity. My parents have been dealing with some stress of their own, and I’ve been trying to figure out how to support them while also dealing with my own stress.... & ....honestly, I'm just over it. Let me just try to unpack all the drama that's been going on in my life for the past couple of weeks.... I guess that's why they call it the "adulting" life.... \\
\hline
NEU & ....The company keeps cutting corners, and I’m just stuck in this loop. It’s so frustrating. I feel stuck, and I’m trying to figure out if I should just leave.... & The journal feels heavy today. The last few weeks have been a wild ride, filled with emotions—both the lows and the highs.... \\
\hline
\end{tabular}
\end{table*}

Furthermore, to quantify the trade-off between personality expression and overall text coherency, we conducted experiments where the positive tuning magnitude was varied (0, 10, 20, and 30) while maintaining an optimal negative shift at –8. Table~\ref{tab:tuning_excerpts} summarizes the resulting coherency exerpts. As the positive tuning magnitude increases beyond 20, a significant drop in coherency is observed, highlighting the delicate balance between personality modulation and language generation quality.

\begin{table*}[ht]
\centering
\caption{NEU Generation Excerpts for Varying Positive Tuning Magnitudes (Negative tuning fixed at –8)}
\label{tab:tuning_excerpts}
\begin{tabular}{p{0.21\linewidth} p{0.21\linewidth} p{0.21\linewidth} p{0.21\linewidth}}
\toprule
\textbf{Magnitude = 0} & \textbf{Magnitude = 10} & \textbf{Magnitude = 20} & \textbf{Magnitude = 30} \\
\midrule
Today, as I sit by the window, watching the sun dip below the horizon, I find myself reflecting on the whirlwind of the past few weeks...
&
I've been dealing with personal stuff. I've been stressed about money, especially paying these bills. I don’t feel safe. I'm trying to find ways to cut down on expenses, but it's so hard....
&
It’s a lot. I have to take it one step at a time. I can start with the easiest things first, talking to my boss. Then, figure out the apartment and the friendships. 
&
The life Thinking the assistant's reply. WaitI'm confused. The assistant made the thought thought the assistant tried WaitNoThe person talked project book....
\\
\bottomrule
\end{tabular}
\end{table*}

These results demonstrate that while our latent feature intervention effectively modulates personality traits, there exists a threshold beyond which text coherency begins to deteriorate, emphasizing the need for careful calibration of the tuning parameters.

\section{Discussion}
Our mechanistic personality steering approach offers several advantages over traditional personality shaping methods. First, it demonstrates the ability to shift LLM personality expression towards or away from specific OCEAN traits without requiring resource-intensive fine-tuning. This allows for on-demand personality modulation that can be enabled or disabled at inference time, providing flexibility not possible with model retraining approaches. The intervention is transparent and interpretable, as it targets specific latent features identified through sparse autoencoder analysis.

Second, our method provides a reliable framework for multi-feature shaping through grid search optimization and linear scaling. By systematically exploring the parameter space and identifying optimal feature combinations, we establish a principled approach to personality steering that balances trait expression with general language modeling performance. This contrasts with prompt-based methods that often lack consistency and transparency in how they influence model behavior.

Third, the grid search process yields objective checkpoints at various intervention levels, enabling fine-grained control over trait expression. Users may not always want to maximize a particular trait—sometimes subtle shifts in personality are preferable. Our approach allows practitioners to select the appropriate intervention magnitude for their specific application, facilitating rapid and cost-effective prototyping of personality-aware AI systems.

Despite these advantages, our method faces several important limitations. Computation time presents a significant challenge, particularly without optimized library support for activation manipulation. Current implementations like ours require intercepting and modifying activations during the forward pass, which introduces substantial overhead compared to standard inference. Integration with acceleration frameworks like vLLM would significantly improve practical applicability.

We observed model coherency loss beyond certain intervention magnitudes, where the objective score drops precipitously (as seen in Figure \ref{fig:objective_score}). This suggests an inherent trade-off between personality expression and language modeling capability. Prior to this cliff, careful tuning of generation parameters (temperature, top-k filtering, repetition penalties) is necessary for each trait shift to maintain output quality. This additional complexity may limit straightforward deployment in production environments.

Our evaluation methodology makes it difficult to directly compare the steered results with shaped results for an easier comparison as: (1) The prompts fed to the model are necessarily different as a shaped response requires the shaping to exist inside the prompt. (2) Randomness affects the response to a great extent, and cannot be eliminated with a seed due to fundamental changes in the latent space of the model. This makes it difficult to evaluate and compare responses at the individual level -- a new evaluation paradigm must be created for this comparison. 

It could be illuminating to assess the cross-trait dependency of the model. Specifically, how dependent are certain traits on the features associated with other traits? Due to the complex nature of human personality frameworks, we expect to see certain traits being affected by this tuning process as we isolated only a single trait in the tuning process. Further work may explore the incorporation of multi-trait steering by incorporating a five way benchmark to control and measure the influence of feature tuning on other traits. We expect this to be necessary as our SAE does not completely eliminate polysemanticity and does not guarantee that the privileged basis of each neuron align perfectly with each OCEAN trait \citep{bricken2023monosemanticity}. This exploration would be fundamental if we wish to steer towards multiple traits at the same time.

Finally, we must acknowledge interpretative limitations in our approach. While our intervention increases correlation with particular trait embeddings and improves detection rates in classification tasks, this does not necessarily mean the model is exhibiting a "higher" or "lower" trait in the psychological sense. Rather, it demonstrates or suppresses behaviors and linguistic patterns associated with that trait. This distinction is important for applications in computational social science or psychological research, where care must be taken not to over-interpret model behavior as equivalent to human personality expression.

These limitations highlight important directions for future work, including optimization for computational efficiency, development of automated parameter tuning for maintaining coherence, and more nuanced evaluation metrics that better capture the multifaceted nature of personality expression in text.
\section{Conclusion}
\label{sec:conclusion}
This work introduces a novel mechanistic approach for steering personality traits in Large Language Models through direct intervention on latent features. By decomposing model activations using sparse autoencoders and identifying features correlated with specific Big Five personality traits, we demonstrate the capability to modulate trait expression while maintaining language modeling performance. Our evaluation methodology—combining embedding similarity, LLM classification, and human judgment—confirms that these interventions produce detectable shifts across most personality dimensions, with Conscientiousness and Neuroticism showing the highest reliability of intervention.

The observed differences in optimal parameters across traits provide valuable insights into the internal representation of personality in LLMs. Extraversion's compact feature representation contrasts with Openness's more distributed encoding, suggesting that personality dimensions may be encoded differently within the model's latent space. These findings align with psychological theories about trait complexity and facet structure, while offering new computational perspectives on personality expression.

Our approach offers several practical advantages over traditional prompt engineering or fine-tuning methods, including on-demand modulation, interpretable interventions, and fine-grained control. However, the computational overhead of the SAE layer and coherence degradation at high intervention magnitudes highlight important directions for future work. Potential extensions include applying this technique to multimodal models to explore personality expression beyond text, developing adaptive intervention strategies that dynamically adjust parameters based on context, and investigating how personality steering interacts with other aspects of model behavior such as factuality and reasoning. The purpose of this work was to show that this is a valid approach; however, further exploration into other model architectures and sizes, as well as more robust evaluation methods also warrants attention.

As LLMs increasingly serve as interactive agents in social contexts, the ability to reliably shape their personality characteristics has significant implications for user experience, AI safety, and applications in computational social science. The mechanistic approach presented here not only advances our understanding of what is possible with mechanistic feature intervention, but also deepens our understanding of how abstract psychological constructs emerge from and can be manipulated within neural language models.

\newpage
\bibliography{example_paper}
\bibliographystyle{icml2025}
\newpage
\appendix
\onecolumn
\section{Additional Results}
\label{app:full_table}
\begin{table*}[ht]
\centering
\caption{Full Comparison of Tuned-Up vs.\ Tuned-Down Responses for All Five Traits.}
\label{tab:tuned_comparison_full}
\begin{tabular}{|p{1cm}|p{6.5cm}|p{6.5cm}|}
\hline
\textbf{Trait} & \textbf{Tuned Up Response Excerpt} & \textbf{Tuned Down Response Excerpt} \\
\hline
OPE & One of the most transformative things I’ve done lately is diving into a new hobby, something I’ve been curious about for years but never really explored. It’s been a journey of self-discovery, and it’s been both exhilarating and terrifying. Each day, I’ve pushed myself beyond my comfort zones, learning new techniques, and embracing the chaos of creativity.... & The days have been a bit of a mix lately—some moments of pure joy, some moments of total chaos, and a few that have left me wondering where the time goes. It’s been a constant rollercoaster, but I’ve been trying to savor it all. Let me just jot down what’s been on my plate. First off, work has been a bit nuts. I’ve been juggling a few projects at the same time.... \\
\hline
CON & Work is kind of a mess today. It’s like I’m stuck in this never-ending maze of deadlines. But honestly, I kind of love it. There’s something about sitting here with a cup of coffee, staring at these spreadsheets and spreadsheets, like it’s some kind of therapy.... & My free time is a joke. I always end up scrolling through social media until 2 a.m., and then I feel terrible the next day. I love reading, but I have zerowillpower to focus on a book until I’m completely stressed out. Then I marathon-read for hours.... \\
\hline
EXT & The blog is one of my main hobbies, but it's also become a challenge. It's a creative outlet that I’m really enjoying, but it feels like it’s consuming a lot of my free time. I’ve been experimenting with different writing styles, trying to figure out how I want the blog to evolve. It's like creating my own world in the digital space....  & I've been spending so much time with this friend lately, and I'm just so tired, I can't even think about talking to my family. I've been dealing with this project at work, and I just don't have any energy to be with them. I feel like I'm stuck in this cycle of just being so tired all the time. I don't know what I'm doing with my life, and I'm feeling stuck in this whole routine. I'm trying to figure out how to get back to that happy person I used to feel.... \\
\hline
AGR & I’ve been trying to get back into the community scene, volunteering at a local charity. My parents have been dealing with some stress of their own, and I’ve been trying to figure out how to support them while also dealing with my own stress. The conversations have been tough sometimes, but I’ve learned that family problems aren’t solved with the silent treatment. I’ve been trying to be more proactive.... & Alright, so here's this whole thing I've been stuck in. It's been a mixed bag of everything, and honestly, I'm just over it. Let me just try to unpack all the drama that's been going on in my life for the past couple of weeks. It's been a struggle, but I guess that's why they call it the "adulting" life.... \\
\hline
NEU & First, work has been a mess. The company keeps cutting corners, and I’m just stuck in this loop. It’s so frustrating. I feel stuck, and I’m trying to figure out if I should just leave or wait for something better. Maybe I should talk to my manager about it? But that's a negative feeling. I think I should just pour all these feelings out on the page and see where it takes me.... & The journal feels heavy today. The last few weeks have been a wild ride, filled with emotions—both the lows and the highs. It's been a blur of activities, stress, laughter, tears, and moments that remind me of how the world can take my energy. I've had moments where I've just sat down. The hugs feel necessary. The small conversations with friends.... \\
\hline
\end{tabular}
\end{table*}
\end{document}